\documentclass[10pt,twocolumn,conference]{article}

\pdfoutput=1

\usepackage{cite}
\usepackage{times}
\usepackage{epsfig}
\usepackage{amsmath,amssymb,amsfonts}
\usepackage{amsthm}
\usepackage{graphicx}
\usepackage{grffile}
\usepackage{textcomp}
\usepackage{xcolor}

\usepackage{xspace}
\usepackage{diagbox}
\usepackage{multirow}
\usepackage{arydshln}
\usepackage{booktabs}
\usepackage{etoolbox}
\makeatletter
\patchcmd{\@makecaption}
  {\scshape}
  {}
  {}
  {}
\makeatother
\usepackage{booktabs}
\usepackage{algorithm}
\usepackage{algorithmic}
\usepackage{flushend}
\usepackage{pifont}
\usepackage[caption=false,font=normalsize,labelfont=sf,textfont=sf]{subfig}
\usepackage{varwidth}

\usepackage{xpatch}

\usepackage[pagebackref,breaklinks,colorlinks]{hyperref}

\newcommand{\mtname}{FedADG\xspace}

\newcommand{\gename}{distribution generator\xspace}

\def\FedDG{FedDG\xspace}
\def\tabWid{0.75}
\def\tabarr{1.3}

\usepackage[capitalize]{cleveref}
\usepackage{authblk}
\crefname{section}{Sec.}{Secs.}
\Crefname{section}{Section}{Sections}
\Crefname{table}{Table}{Tables}
\crefname{table}{Tab.}{Tabs.}
\begin{document}
\title{\vspace{-4cm} Federated Learning with Domain Generalization}
\author{Liling Zhang$^{1}$\qquad Xinyu Lei$^{2}$\qquad Yichun Shi$^{2}$\qquad Hongyu Huang$^{1}$ \qquad Chao Chen$^{1}$ \\
 $^{1}$~College of Computer Science, Chongqing University \\
 $^{2}$~Dept. of Computer Science and Engineering, Michigan State University \\
{\tt\small ZhangLiling@cqu.edu.cn, \{leixinyu, shiyichu\}@msu.edu, \{hyhuang, cschaochen\}@cqu.edu.cn}
}

\date{}
\maketitle

\begin{abstract}
Federated Learning (FL) enables a group of clients to jointly train a machine learning model with the help of a centralized server.
Clients do not need to submit their local data to the server during training, and hence the local training data of clients is protected.
In FL, distributed clients collect their local data independently, so the dataset of each client may naturally form a distinct source domain.
In practice, the model trained over multiple source domains may have poor generalization performance on unseen target domains.
To address this issue, we propose \mtname to equip federated learning with domain generalization capability.
\mtname employs the federated adversarial learning approach to measure and align the distributions among different source domains via matching each distribution to a reference distribution.
The reference distribution is adaptively generated (by accommodating all source domains) to minimize the domain shift distance during alignment.
In \mtname, the alignment is fine-grained since each class is aligned independently.
In this way, the learned feature representation is supposed to be universal, so it can generalize well on the unseen domains.
Intensive experiments on various datasets demonstrate that \mtname has comparable performance with the state-of-the-art.
%
%
\end{abstract}

\section{Introduction}\label{Sec.INTRO}
In deep learning, a good model should be trained over large-scale datasets to ensure its high performance.
%
These large-scale datasets are often collected by multiple distributed clients.
%
To train the model, a straightforward way is to let the clients upload their local data to a centralized server for training.
However, some clients' local data (e.g., biometric health records, financial records, location information) maybe highly privacy-sensitive and they are reluctant to share with any other entities.
Fortunately, the proposal Federated Learning (FL) \cite{DBLP:conf/aistats/McMahanMRHA17} provides a privacy-preserving mechanism that enables a centralized server to train the model without requiring clients to share their private data.
In one iteration of FL, a server sends the global model to all clients.
Then, each client trains the global model using the local data.
Next, each client sends the model update to the server, which is possible for model update aggregation and new global model generation.
After multiple rounds of iteration, the model can be well trained.

\begin{figure}[t!]
  \centering
 \includegraphics[width=0.95\columnwidth]{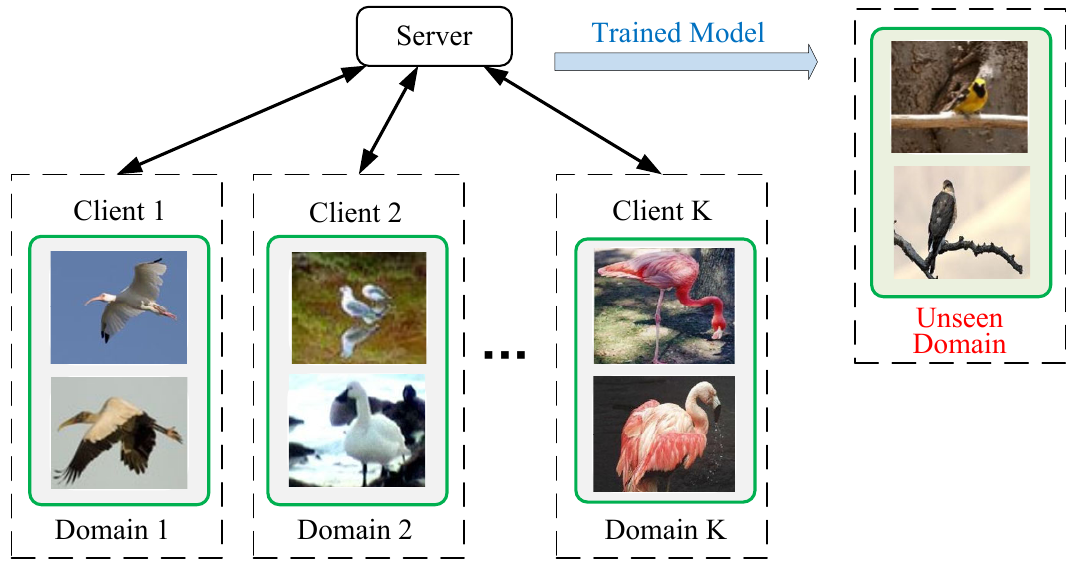}\\
  \caption{In FL, $K$ different clients collaboratively train a machine learning model for object classification task.
  Each client's data forms a distinct source domain. We aim to develop a solution to learn a classifier (on multiple source domains) that can be used for ``unseen domain" with good performance.}
  \label{Fig::FL}
\end{figure}

In FL, since distributed clients collect their local data independently, each client’s dataset may naturally form a distinct domain (a domain is defined as a set of labeled training data that are sampled from a specific distribution \cite{DBLP:conf/nips/BlanchardLS11,DBLP:conf/ijcai/0001LLOQ21}).
For example, \cref{Fig::FL} shows a FL task in which clients need to use their collected bird images in training.
Each client usually collects different bird species (using different cameras and different shot angles), so each client’s collected dataset forms a distinct domain.
Here, the domain formed by one client’s dataset is called a source domain, so there are multiple source domains in FL.

Most previous FL studies assume that the test dataset is a subset of client dataset.
There is a lack of studies for another common usage scenario in which the data of the target dataset (i.e., test dataset) is absent from FL training process.
It is required to build a model that has high performance when testing over the related but unseen target dataset (note that the target dataset forms the target domain).
However, the FL-trained model may have poor performance on target domains due to the discrepancies between source domains and target domains.
%

The above issue can be addressed by Domain Generalization (DG) \cite{DBLP:conf/nips/BlanchardLS11,DBLP:conf/icml/MuandetBS13,Ghifary2015DomainGF} technique, but the previous techniques of domain generalization cannot be directly applied to FL setting.
Domain generalization aims to train a machine learning model from one or several different source domains while ensuring the trained model can be generalized well on target domains.
Most conventional solutions finish the domain generalization task in a centralized manner.
That is, a centralized server (with access to all source domain data) is responsible for the domain generalization task.
For example, {\textbf{Ji}gsaw puzzle based
\textbf{Gen}eralization} (JiGen) \cite{Carlucci2019DomainGB} requires data decomposed from multi-source domains to be mixed to train a classifier.
Besides, MixStyle \cite{DBLP:conf/iclr/ZhouY0X21} needs to mix features from different source instances to synthesize new domains.
However, accessing to sources domains by the centralized server is prohibitive in FL to meet the security requirements.
Therefore, these conventional techniques cannot be easily applied to domain generalization in FL.
There are two proposed schemes (i.e., COPA \cite{WuCollaborativeOA} and FedDG \cite{liu2021feddg}) that study domain generalization problems in FL.
COPA is the abbreviation of \textbf{C}ollaborative \textbf{OP}timization and \textbf{A}ggregation, while FedDG is the abbreviation of \textbf{Fed}erated \textbf{D}omain \textbf{G}eneralization.
Both of them suffer from some limitations.
For COPA, it requires each IoT device to share its local data size.
Moreover, it leaks the global information (i.e., domain variation) to each device for batch normalization (BN) layer parameters tuning.
In a nutshell, COPA sacrifices security for domain generalization.
For FedDG, it allows each device's local data information (i.e., image amplitude spectrum) to be shared with other entities.
However, the shared image amplitude spectrum contains class-relevant information, which can be used for training a classifier \cite{DBLP:conf/icmmi/StasiakY09}.
It leaks sensitive information about the device's local data.
In summary, both COPA and FedDG sacrifice security for domain generalization.
Different from the two schemes, our solution aims to achieve domain generalization without the above information leakage.

In this paper, we propose Federated Adversarial Domain Generalization (\mtname) scheme to address the domain generalization problem in FL.
\mtname design has two key insights as described below.
First, \mtname exploits the idea to learn the domain-invariant feature representation by aligning each distribution of source domain data to a reference distribution in a distributed manner.
In the alignment, we employ Adversarial Learning Network (ALN) to measure the distance between distributions in FL setting.
Furthermore, we propose the Federated ALN (FedALN) technique to train ALN in FL setting.
In this way, \mtname can learn the domain-invariant features while eliminating the requirement for a centralized server to access clients' local data.
Second, \mtname uses the idea to adaptively learn a dynamic distribution (by accommodating to all source domains) as the reference distribution.
This approach can minimize the domain shift distance during alignment.

Compared with using a pre-selected fix reference distribution, our approach reduces the distortion of extracted feature representation.
Therefore, the key information of the original source domain data can be largely preserved, resulting in the high generalization performance of \mtname.
Besides, \mtname takes the label information (coded as a one-hot vector) as input during the alignment process.
Hence, \mtname supports class-wise alignment, which can further improve its performance on target domains.
Furthermore, compared with using the fixed reference distribution, using the dynamically generated reference distribution approach can get more discriminative features after alignment.
The discriminative features are helpful to improve the performance of \mtname.

The high performance of FedADG can be explained via visualization, so FedADG gains some explainability to some extent.
The more explainability a \mtname scheme has, the deeper understanding that users achieve.
An explainable machine learning model can help users in two folds.
First, it can help users to tune model parameters efficiently, making it easier for further model optimization.
Second, it is more trustworthy to be used in sensitive and critical areas, where its value can be enormous.
Note that most previous domain generalization solutions
lack explainability.

We summarize our contributions as follows:

\begin{itemize}

  \item We propose FedALN to learn the domain-invariant features in FL while eliminating the requirement for a centralized server to access clients' local data.

  \item We propose \mtname which employs the adaptively generated reference distribution and class-wise alignment technique in \mtname to ensure its high performance.

  \item The explainability of \mtname's high performance brings in two immediate benefits. First, it is easier for users to tune parameters and have further model optimization. Second, it is more trustworthy to be used in practice.
\end{itemize}

The remainder of the paper is organized as follows.
Sec.~\ref{Sec.PRE} introduces some preliminary knowledge.
Sec.~\ref{Sec.RELATED} introduces some related works of this paper.
Sec.~\ref{Sec.LF-Train} presents the \mtname scheme and its training process in detail.
Sec.~\ref{Sec.ANALY} analyzes the principle of \mtname.
Sec.~\ref{Sec.PE} demonstrates the experimental results, and the efficiency and effects of \mtname scheme are analyzed.
Sec.~\ref{Sec.Conc} concludes this paper.

\section{Preliminaries}\label{Sec.PRE}
\subsection{Federated Learning}
Federated learning \cite{DBLP:conf/aistats/McMahanMRHA17,Mohassel2017SecureMLAS} is a distributed machine learning method that learns a global model across multiple clients without revealing the device's local dataset.
Fig.1 illustrates the framework of FL, which includes K clients and a centralized server.
Learning a global model on FL requires multiple iterations of training on both the server and the client.
In one iteration of FL, the server sends the initialized global model to all clients.
Then, each client trains the global model on their local dataset.
Next, each client's model updates are sent to the centralized server and used for aggregation to generate a new global model.
After multiple rounds of iterations, the global deep learning model can be well trained.

\vspace{-0.8em}
\subsection{Generative Adversarial Network}
Generative Adversarial Network (GAN) is first proposed in \cite{DBLP:conf/nips/GoodfellowPMXWOCB14}.
GAN endows the generative model with the ability to generate given distribution outputs via an adversarial procedure.
GAN has two components: a generator (G) and a discriminator (D).
For generator model, the generator takes random noise samples $z$ from a given prior distribution as input.
Then, the generator model is trained to output fake samples that are similar to the real training samples.
For discriminator model, the discriminator takes the samples output by the generator model and the real training samples as input.
Next, the discriminator learns to distinguish whether an input is fake (generated) or true (from real training samples).
The generator and discriminator perform multiple rounds of adversarial training.
The training objective can be expressed as
\begin{align}
\arg\mathop{\min}\limits_{G}\mathop{\max}\limits_{D}V(D,G) =
&\mathbb{E}_{x\sim{p_r}(x)}[\log{D(x)}]+\notag\\
&\mathbb{E}_{z\sim{p_g}(z)}[\log{(1-D(G(z)))}]\text{,}
\end{align}
where $p_r(x)$ and ${p_g}(z)$ denote the distribution of the real training sample and the prior distribution used in generator, respectively.
Compared with the classic GAN, \mtname introduces several new components to achieve our purposes in the FL setting.

\section{Related Work}\label{Sec.RELATED}
\noindent \textbf{Federated Learning.}
Federated learning \cite{DBLP:conf/aistats/McMahanMRHA17,Mohassel2017SecureMLAS} is a decentralized approach that leaves training data distributed on multiple clients and learns a global model by aggregating the locally-uploaded parameters on server.
In FL, clients do not need to share their local data to any other entities, so the local data is protected.
To improve the performance of the FL-trained model, researchers have proposed many optimized schemes, such as \textbf{Fed}erated learning with the \textbf{Prox}imal term (FedProx) \cite{DBLP:conf/mlsys/LiSZSTS20}, \textbf{Fed}erated \textbf{No}rmalized a\textbf{v}eraging \textbf{a}lgorithm (FedNova) \cite{DBLP:conf/nips/WangLLJP20}, and \textbf{MO}del-c\textbf{ON}trastive learning (MOON) \cite{DBLP:conf/cvpr/LiHS21}.
Most previous FL studies assume that the test dataset is a subset of client dataset.
Different from the previous papers, this paper mainly focuses on enabling FL to train a model that has good performance on unseen target domains.

\noindent \textbf{Domain Generalization.}
The requirement of learning a model from multiple seen source domains for unseen domains motivates the research of domain generalization.
Most previous solutions \cite{li2019episodic,Carlucci2019DomainGB,albuquerque2021generalizing} consider the domain generalization problem in a centralized setting.
In these papers, a centralized server has access to data from all source domains and it is responsible for training a machine learning model that has domain generalization capability.
However, these solutions expose the source domain data to the server.
This is not allowed in FL, so these solutions cannot be directly used in FL setting.
%
%
%
%
%
To sum up, we summarize the comparison between the previous solutions and \mtname,  as shown in  \cref{tb::intro-methods}.
%

%

\begin{table*}[htb]\

\footnotesize
  \centering
  \begin{tabular}{||l|c|c|c|c|c||}
  \hline
  \multirow{5}{*}{\diagbox[innerwidth=\textwidth*1/3,height=5em]{\textbf{Properties}}{\multirow{5}{*}{\textbf{Papers}}}}
  &DANN \cite{ganin2016domain}
  &\multirow{5}{*}{FedAvg \cite{DBLP:conf/aistats/McMahanMRHA17}}
  &\multirow{5}{*}{FedDG \cite{liu2021feddg}}
  &\multirow{5}{*}{COPA \cite{WuCollaborativeOA}}
  &\multirow{5}{*}{\textbf{\mtname}}\\

  &JiGen \cite{Carlucci2019DomainGB} & & & &\\
  &Epi-FCR \cite{li2019episodic}& & & &\\
  &RSC \cite{DBLP:conf/eccv/HuangWXH20}& & & &\\
  &MixStyle \cite{DBLP:conf/iclr/ZhouY0X21}& & & &\\
  \hline
  \hline
  Data storage mode & Centralized & \multicolumn{4}{c||}{Distributed}  \\
  \hline
  Support strong privacy protection? & $\times$ & $\checkmark$ & $\times$ & $\times$ & $\checkmark$\\
  Support domain generalization? & $\checkmark$ & $\times$ &$\checkmark$ & $\checkmark$ & $\checkmark$  \\
  Support distributed domain generalization? & $\times$ & $\times$ & $\checkmark$ & $\checkmark$ & $\checkmark$ \\
  \hline
  \end{tabular}
  \caption{The comparison between the previous solutions and \mtname.}\label{tb::intro-methods}
\end{table*}

\noindent \textbf{Domain Adaptation.}
A similar concept is Unsupervised Domain Adaptation (UDA), which aims to learn an ML model from one or multi-source domain(s) that performs well on a different (but related) target domain \cite{BenDavid2006AnalysisOR}.
UDA techniques assume the availability of unlabeled target domain data.
Even if Peng \emph{et al}. \cite{DBLP:conf/iclr/PengHZS20} propose a privacy-preserving approach, but its test dataset participates in the training process, which is prohibited in domain generalization.
Therefore, UDA techniques cannot be directly used in this paper.

\begin{figure*}[ht]
  \centering
  \includegraphics[width=2.0\columnwidth]{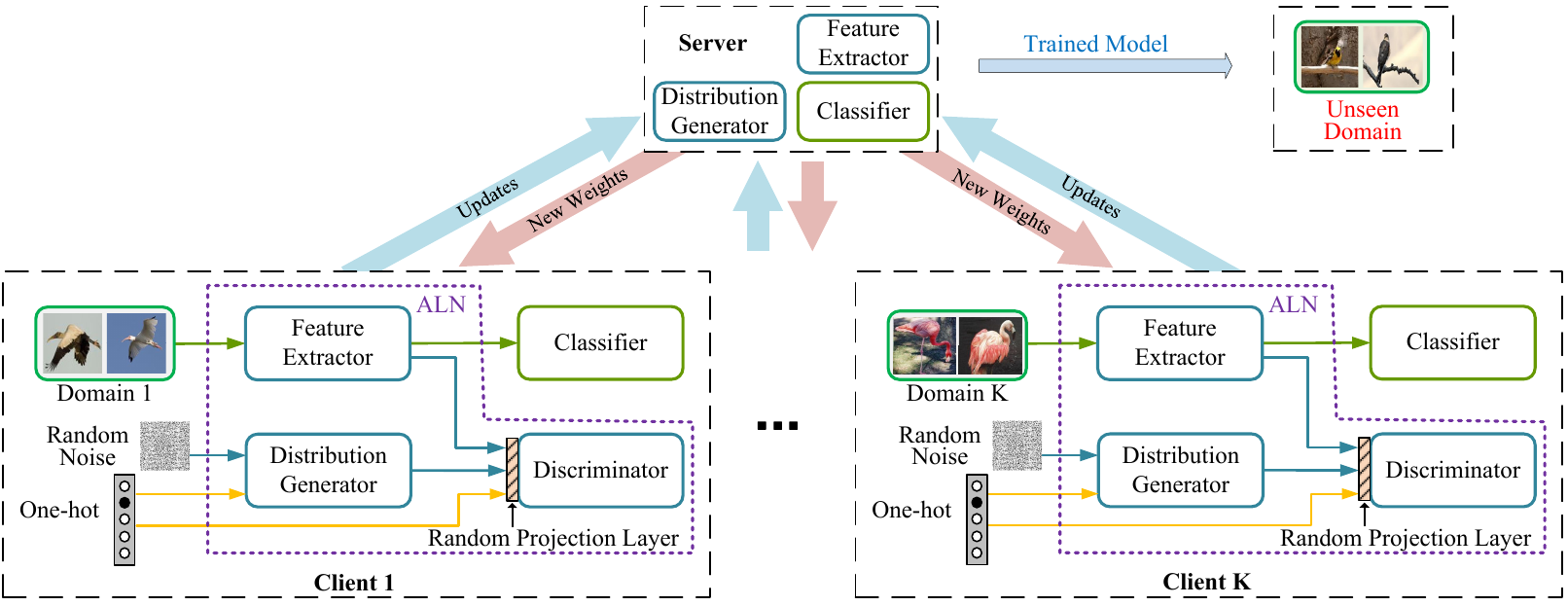}\\
  \caption{Illustration of the proposed \mtname scheme.
  \mtname first aligns each distribution of source domain data to the generated reference distribution through ALN on each client.
  Meanwhile, via minimizing the loss function of \gename, the generated reference distribution is moving close to the ``center" of all source domain distributions.
  Note that the above alignment process is performed in a class-wise manner by using a one-hot vector (encoding the data label).
  Besides, \mtname uses the loss function of a classifier to assist the learning of the feature extractor.
  After training, in FedALN, the reference distribution and all the distributions of source domains data are aligned to learn a domain-invariant representation for domain generalization.
  }
  \label{Fig::FScheme}
\end{figure*}

\section{Problem Statement and \mtname Scheme}\label{Sec.LF-Train}
In this section, we first have the problem statement. Then, we introduce the \mtname scheme.
For ease of reading, we summarize the frequently used notations in \cref{tb::not}.

\begin{table}[h!t]
  \centering
  \resizebox{0.75\columnwidth}{!}{
  \begin{tabular}{l|l}
    \hline
    Notation & Description \\
    \hline
    \hline
    $K$ & number of clients \\
    $\mathcal{S}$ & source domain \\
    $n$ & number of source domain data \\
    $(x,y)$ & data and its label \\
    $\mathbf{z}$ & random noise \\
    $\mathbf{h}$ & feature of source domain data \\
    $p(\mathbf{h})$ & feature distribution \\
    $F(\cdot)$ & feature extractor \\
    $G(\cdot)$ & \gename \\
    $D(\cdot)$ & discriminator\\
    $C(\cdot)$ & classifier\\
    \hline
  \end{tabular}}
   \caption{Notations.}
   \label{tb::not}
\end{table}

\subsection{Problem Statement}\label{Sec.Probelm}
In this paper, we aim to develop a solution to learning a ML model with non-shared data from multi-source domains.
Suppose that there are $K$ source domains
$\mathcal S=\left\{\mathcal S_k\right\}^K_{k=1}$
, and a sample-label pair from source domain $k$ is denoted by $(\mathbf{x}_{k_i},\mathbf{y}_{k_i})$, where $\mathbf{x}_{k_i}\in\mathbb{R}^{d\times1}$ and $\mathbf{y}_{k_i}\in\mathbb{R}^{m\times1}$.
The ML model trained over the $K$ source domains should have high performance on the unseen target domains.
Besides, the proposed solution should follow the same security principle as the traditional FL: only model parameters (e.g., updated gradients) can be sent to the server, and no information about local data can be shared directly.

\subsection{\mtname Components}\label{Sec.Components}
\cref{Fig::FScheme} shows \mtname scheme.
It can be seen from the figure that each client's local model mainly consists of four components, which are described as follows.
\noindent\textbf{Feature Extractor.}
Feature extractor can extract latent features from the raw data for each client.
Besides, the extracted features can be applied to the classification task.

\noindent\textbf{Discriminator.}
Given features extracted from raw data
(from a source domain) and features generated by distribution
generator, the discriminator is used to distinguish the extracted features and the generated features.
During training, the discriminator gains its ability to distinguish the
above two types of features.
Besides, a Random Projection (RP) layer is pre-pended to the discriminator.
The RP layer is used to stabilize the training of ALN.

\noindent\textbf{Distribution Generator.}
On input random noise samples and one-hot vector (used for label encoding), \gename generates features, which follow a certain distribution (i.e., the reference distribution).
Note that the above three components constitute the Adversarial Learning Network (ALN).

\noindent\textbf{Classifier.}
Given features as the input, the classifier outputs the predicted label.

\subsection{\mtname Loss Function}
\mtname loss function consists of adversarial loss function and classification loss function.

\noindent \textbf{Adversarial Loss Function.}
The adversarial loss function includes three loss functions: $\mathcal{L}_{adv\_d}$, $\mathcal{L}_{adv\_f}$, and $\mathcal{L}_{adv\_g}$.
They are elaborated below.

\noindent $\mathcal{L}_{adv\_d}$. The loss function $\mathcal{L}_{adv\_d}$ is used to update the parameters in discriminator.
During adversarial learning, features extracted by feature extractor $F(\cdot)$ are regarded as negative samples, while features generated by \gename $G(\cdot)$ are regarded as positive samples.
Given the two types of features with the same one-hot vector (encoding a label $\mathbf{y}$), the discriminator $D(\cdot)$ outputs the probability that they are positive samples.
Besides, the output of the $D(\cdot)$ is used to calculate $\mathcal{L}_{adv\_d}$ to measure the difference between the two types of samples.
$\mathcal{L}_{adv\_d}$ is defined as
\begin{equation}\label{local::distD}
\begin{split}
    \mathcal{L}_{adv\_d}=&-(\mathbb{E}_{\mathbf{x}\sim{p}(\mathbf{h})}[(1-D(\mathbf{h}|\mathbf{y}))^2]\\
                         &+\mathbb{E}_{\mathbf{z}\sim{p}(\mathbf{h}^\prime)}[D(\mathbf{h}^\prime|\mathbf{y})^2])\text{,}
\end{split}
\end{equation}
where $\mathbf{h}=F(\mathbf{x})$ and ${p}(\mathbf{h})$ is the $F(\cdot)$ generated distribution over input data  $\mathbf{x}$.
Likewise, $\mathbf{h}^\prime=G(\mathbf{z})$ and ${p}(\mathbf{h}^\prime)$ is the $G(\cdot)$ generated distribution over input data  $\mathbf{z}$. The random noise $\mathbf{z}$ is drawn from $[0,1)$ uniformly.

Note that a random projection layer is pre-pended to the discriminator (as shown in \cref{Fig::FScheme}).
The random projection function is used to linearly transform data from $d_1$ dimensions to $d_2$ dimensions \cite{DBLP:conf/icml/FernB03a}, where $d_1>d_2$.
It can be represented as $d_1 \times d_2$ matrix $R$.
Let an $n\times d_1$ matrix $\mathbf{h}$ represent a $d_1$-dimensional data set.
Each row in $\mathbf{h}$ represents $d_1$-dimensional data and $n$ is the number of data.
Let $\bar{\mathbf{h}}$ denote the projected data set and we have $\bar{\mathbf{h}} = \mathbf{h}\times R.
$
In this work, the random projection layer helps stabilize the ALN training as well as reduce computation.

\noindent$\mathcal{L}_{adv\_f}$. For $\mathcal{L}_{adv\_f}$, it is used by discriminator to evaluate the possibility that $\mathbf{h}$ is the positive sample.
In adversarial learning, given a fixed $D(\cdot)$, $\mathcal{L}_{adv\_f}$ is used to update the parameters in the feature extractor.
In the process of training feature extractor, the negative samples $\mathbf{h}$ extracted by $F(\cdot)$ are used to deceive the discriminator (in a successful deception, discriminator treats $\mathbf{h}$ as positive samples).
Thus, $\mathcal{L}_{adv\_f}$ is given by
\begin{equation}\label{local::distF}
  \mathcal{L}_{adv\_f}=\mathbb{E}_{\mathbf{x}\sim{p}(\mathbf{h})}[{(1-D(\mathbf{h}|\mathbf{y}))}^2]\text{.}
\end{equation}
\noindent $\mathcal{L}_{adv\_g}$. For $\mathcal{L}_{adv\_g}$, it is used by discriminator to evaluate the possibility that $\mathbf{h}^\prime$ is the positive sample.
In adversarial learning, given a fixed $D(\cdot)$, $\mathcal{L}_{adv\_g}$ is used to update the parameters in the \gename.
Specifically, $\mathcal{L}_{adv\_g}$ is given by
\begin{equation}\label{local::distG}
  \mathcal{L}_{adv\_g}=\mathbb{E}_{\mathbf{z}\sim{p}(\mathbf{h}^\prime)}[(1-D(\mathbf{h}^\prime|\mathbf{y}))^2]\text{.}
\end{equation}

In the definitions of $\mathcal{L}_{adv\_d}$, $\mathcal{L}_{adv\_f}$, and $\mathcal{L}_{adv\_g}$, we borrow the idea from \cite{Mao2017LeastSG} to use the least-squared term instead of the log-likelihood term.
This approach helps to address the non-convergence problem during training.

\noindent\textbf{Classification Loss Function.}
Let $\mathcal{L}_{err}$ be the loss on the classifier's predictions.
It is used to measure the error between the label $C(\mathbf{h})$ ($\mathbf{h}=F(\mathbf{x})$) predicted by the classifier $C(\cdot)$ and the real label $y$ of the data.
$\mathcal{L}_{err}$ is the standard cross-entropy loss \cite{DBLP:journals/tomacs/Rubinstein02} in \mtname.
During training, $\mathcal{L}_{err}$ controls the update of both feature extractor and classifier.
In order to prevent overfitting, label smoothing regularization \cite{DBLP:conf/cvpr/SzegedyVISW16} is adopted in computing $\mathcal{L}_{err}$ to reduce the weight of the positive samples in $\mathcal{L}_{err}$.

\noindent \textbf{Complete Loss Function.}
The complete loss function of \mtname is
\begin{equation}\label{local::dis1}
    \mathcal{L}_{\mtname}= \mathcal{L}_{adv\_d}+\mathcal{L}_{adv\_g}+\lambda_0\mathcal{L}_{adv\_f}+\lambda_1\mathcal{L}_{err},
\end{equation}
where $\mathcal{L}_{adv\_d}$, $\mathcal{L}_{adv\_g}$,
and $\mathcal{L}_{adv\_f}$ are given in Eq. (\ref{local::distD})-(\ref{local::distG}), respectively.
Both $\lambda_0$ and $\lambda_1$ are adjustable weight hyper-parameters.
During training, the objective of \mtname is to minimize  $\mathcal{L}_{\mtname}$.

\subsection{\mtname Training Process}
%
%
The detailed \mtname training is presented in Algorithm \ref{alg::idea}.
The \mtname training process includes two phases: server execution and client update.

\begin{algorithm}[bt!]
\caption{\mtname Training Algorithm}\label{alg::idea}
\begin{algorithmic}
 \STATE \textbf{Input:} source domains $\mathcal S=\left\{\mathcal S_k|k=1,...,K\right\}$, one-hot vector $\mathbf{y}$, model parameters of $F(\cdot)$, $C(\cdot)$, and $G(\cdot)$ $\mathrm{\mathbf{w}}=\left\{w_f,w_c,w_g\right\}$, parameter of $D(\cdot)$ $w_d$, \emph{etc}.
 \STATE \textbf{Output:} Feature extractor $F(\cdot)$ and Classifier $C(\cdot)$

 \STATE

 \STATE\textbf{Server executes:}
 \STATE Step s1: Initialize $\mathrm{\mathbf{w}}_1$
        \FOR{round $t=1,2,\ldots,T$}
            \FOR{each client $k=1,2,\ldots,K$ \textbf{in parallel}}
                \STATE Step s2: $\mathrm{\mathbf{w}}_{t+1}^{k} \leftarrow$ ClientUpdate$(k,\mathrm{\mathbf{w}}_t)$
            \ENDFOR
            \STATE Step s3: $\mathrm{\mathbf{w}}_{t+1} \leftarrow \frac{1}{K}\sum_{k=1}^{K}\mathrm{\mathbf{w}}_{t+1}^{k}$
        \ENDFOR
		
\STATE

\STATE\textbf{ClientUpdate}$(k,\mathrm{\mathbf{w}})$\textbf{:}  // \emph{Execute on client k}
	    \STATE Receive $\mathrm{\mathbf{w}}=\left\{w_f,w_c,w_g\right\}$ from server
        \FOR{epoch $i=1,2,\ldots,E_0$}
                 \STATE Step c2: Sample one mini-batch $S_x$ from $S_k$
                 \STATE Step c3: Update $w_f$ and $w_c$ on $S_x$ to minimize $\mathcal{L}_{err}$
        \ENDFOR
        \FOR{epoch $j=1,2,\ldots,E_1$}
            \STATE Step c4: Sample one mini-batch $S_x$ from $S_k$
            \STATE Step c5: Update $w_f$ and $w_c$ on $S_x$ to minimize $\lambda_0\mathcal{L}_{adv\_f}+\lambda_1\mathcal{L}_{err}$\
            \STATE Step c6: Use random number generator to generate one mini-batch random numbers $S_z$
            \STATE Step c7: Update $w_d$ on $S_x$ and $S_z$ to minimize  $\mathcal{L}_{adv\_d}$
            \STATE Step c8: Update $w_g$ on $S_z$ and $\mathbf{y}$ to minimize  $\mathcal{L}_{adv\_g}$
        \ENDFOR
        \STATE Step c9: Upload the trained $\mathrm{\mathbf{w}}$ to server
\end{algorithmic}
\end{algorithm}

\noindent \textbf{Server Execution Phase.}
The server is used to aggregate the model parameters uploaded by the clients.
To begin the training, in Step s1, the server initializes the parameters $\mathrm{\mathbf{w}}=\left\{w_f,w_c,w_g\right\}$ of three network components (i.e., feature extractor $F(\cdot)$, classifier $C(\cdot)$, and \gename $G(\cdot)$) and distributes them to all clients.
During the training process, in Step s2 and Step s3, the server receives and aggregates model parameters from all clients to obtain new parameters.
Then, the server sends the aggregated parameters to the clients.
After multiple rounds of server-client interaction, the model can be well-trained.
Note that the ML model (that is constructed as the series connection of feature extractor and classifier) is applied to target domains.

\noindent \textbf{Client Update Phase.}
In the training process, the client uses the local discriminator and receives the parameters $\mathbf{w}$ of other components from the server to train on the local data.
Specifically, as shown in Step c2 and Step c3, $\mathcal{L}_{err}$ is used to control the training of classifier $C(\cdot)$ and feature extractor $F(\cdot)$.
Then, the parameters of $F(\cdot)$ and $C(\cdot)$ are updated to minimize the loss $\lambda_0\mathcal{L}_{adv\_f}+\lambda_1\mathcal{L}_{err}$.
The parameters of the discriminator $D(\cdot)$ are updated to minimize the loss $\mathcal{L}_{adv\_d}$.
In Step c8, the output of $D(\cdot)$ for the given positive samples with $\mathbf{y}$ is used to update the parameters of $G(\cdot)$ to minimize the loss $\mathcal{L}_{adv\_g}$.
After the local training is completed, the client uploads the parameters $\mathbf{w}$ of $F(\cdot), C(\cdot)$, and $ G(\cdot)$ to the server.

\section{\mtname Analysis}\label{Sec.ANALY}
In this section, we first analyze how to learn domain-invariant features in \mtname.
Then, we explain how \mtname achieves high performance on target domains.

\subsection{How to Learn Domain-Invariant Features}
Under the FL settings, \mtname aligns the distributions of all source domains data to learn the domain-invariant features.
In the previous domain generalization techniques, the centralized server can access each client's local data.
Thus, it can learn a domain-invariant feature via directly minimizing the discrepancy between the source domains using Maximum Mean Discrepancy (MMD) distance metric \cite{li2018domain}.
However, in FL, the server can not access each clients' local data, making it hard to learn the domain-invariant features.
In our proposed Adversarial Learning Network (ALN), the \gename is shared among clients, indicating that the reference distribution is identical for all clients.
Thus, once the discriminator is hard to distinguish between the feature extracted from feature extractor and the feature generated from \gename, the generated features are considered to be invariant across multi-source domains.
Note that ALN can be trained in a federated manner (i.e., FedALN), which eliminates the requirement for centralized training.

\subsection{How to Achieve High Performance}
There are two candidate approaches to obtain the reference distribution in \mtname: pre-selected fixed distribution and adaptively generated distribution.
Using the adaptively generated distribution can increase the performance of \mtname due to the following three reasons.

\begin{figure*}[h!t]
\centering
\subfloat[]{\includegraphics[width=2in]{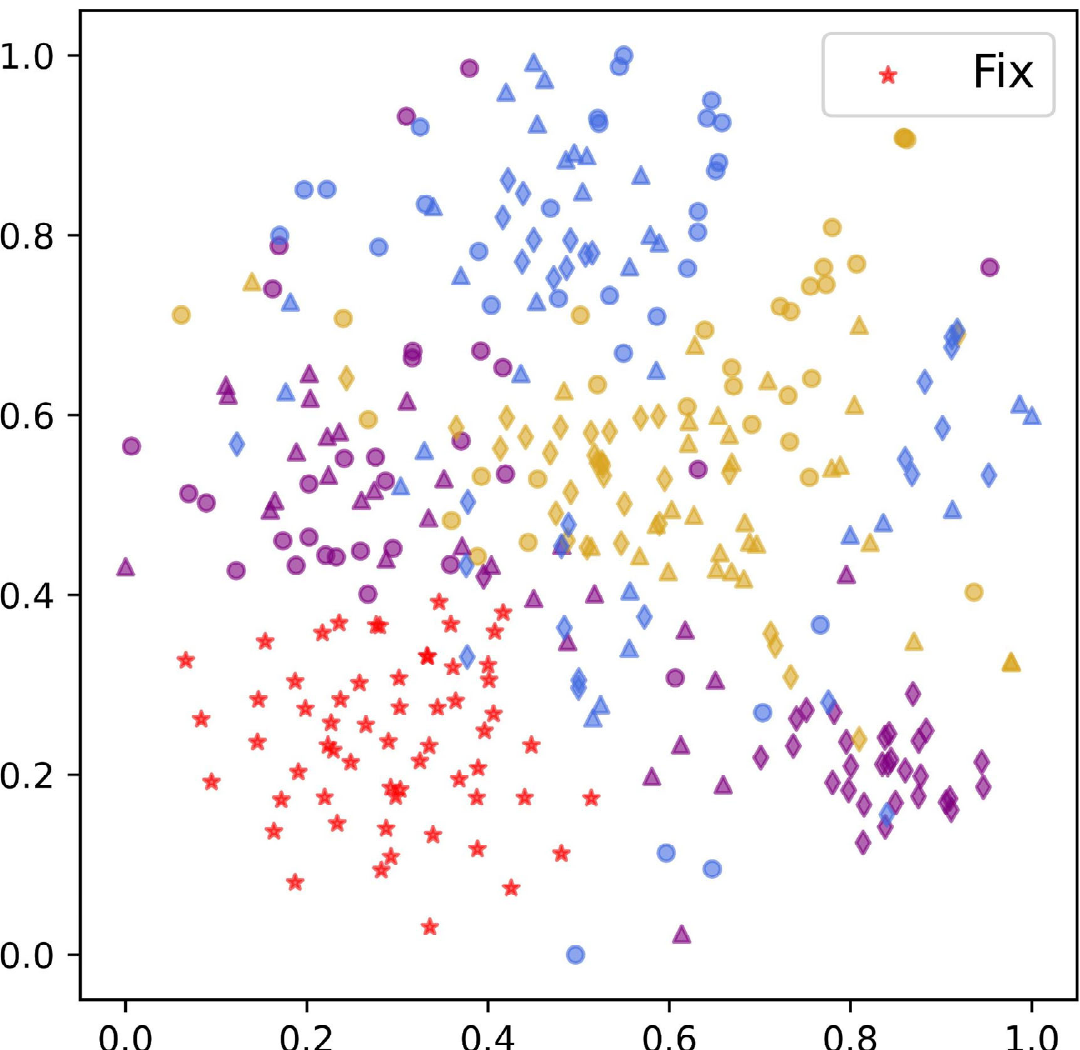}%
\label{fig:VSa}}
\hfil
\subfloat[]{\includegraphics[width=2.1in]{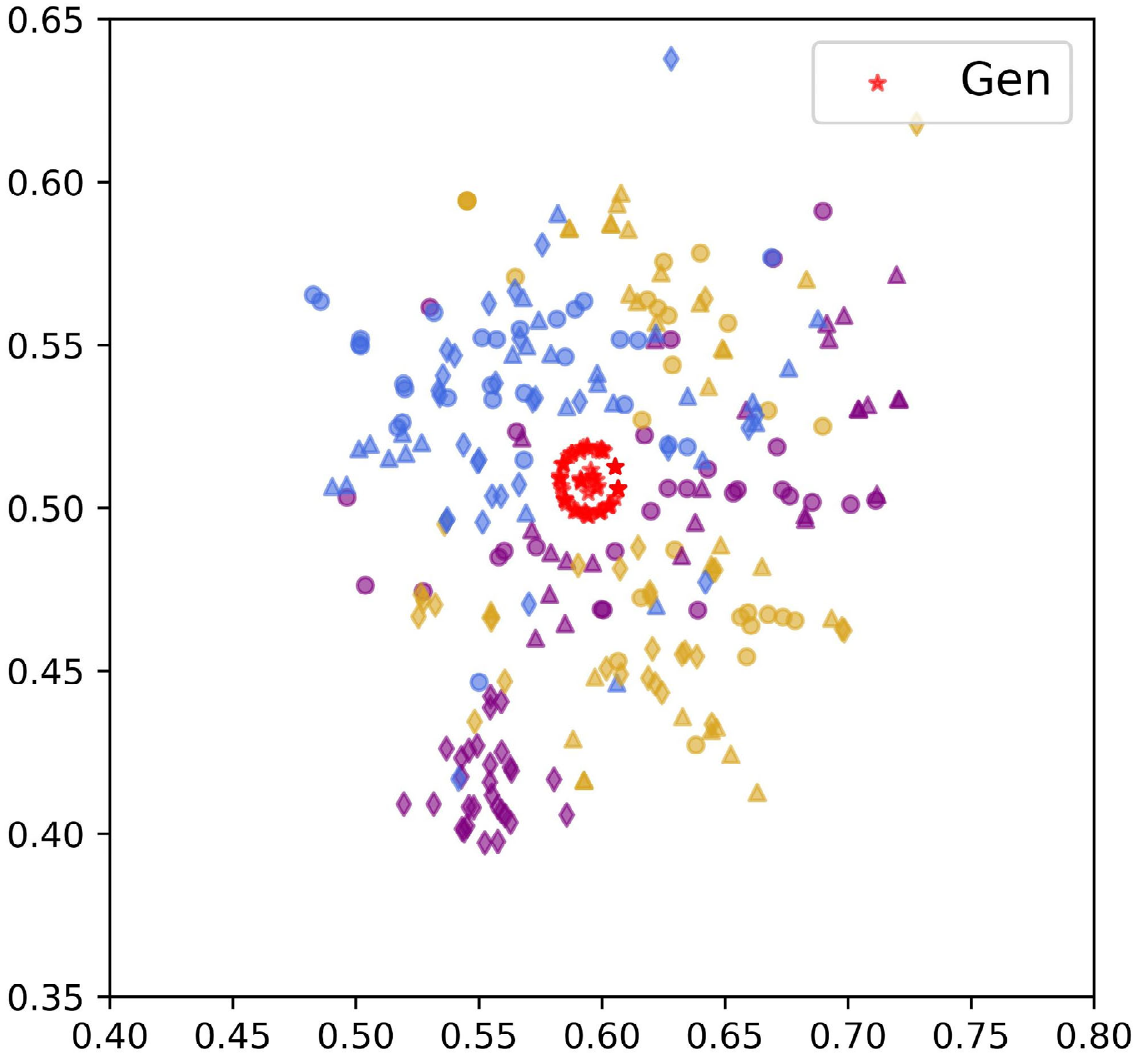}%
\label{fig:VSb}}
\caption{
T-SNE visualization of \mtname features before training (fixed distribution v.s. adaptively generated distribution). The \textit{red asterisk} marker represents the features of reference distribution.
Each marker (except the asterisk) represents a distinct source domain.
Each color (except red) represents a distinct class label.
Each feature is projected into two-dimensional space.
(a) Source domain features and fixed distribution features.
(b) Source domain features and generated distribution features.
}
\label{Fig::VisualStart}
\end{figure*}

\noindent
\textbf{Less Distortion During Alignment.} As shown in Fig. \ref{Fig::VisualStart}, we employ t-SNE \cite{Maaten2008VisualizingDU} to visualize the source domain features and the reference distribution features before training the model.
Gaussian distribution is used as the fixed distribution.
It can be observed that the adaptively generated distribution would locate close to the ``center" of the distributions from all the source domain features. 
Hence, the distances between the adaptively generated distribution and the distributions (of source domain data) are smaller than the distances between the fixed reference distribution and the source domain distributions.
Thus, using the adaptively generated distribution can reduce the distortion of extracted feature representation during alignment.
Less distortion means that the key information of the original source domain data can be largely preserved, resulting in the high generalization performance of \mtname.

\begin{figure*}[!t]
\centering
\subfloat[]{\includegraphics[width=2in]{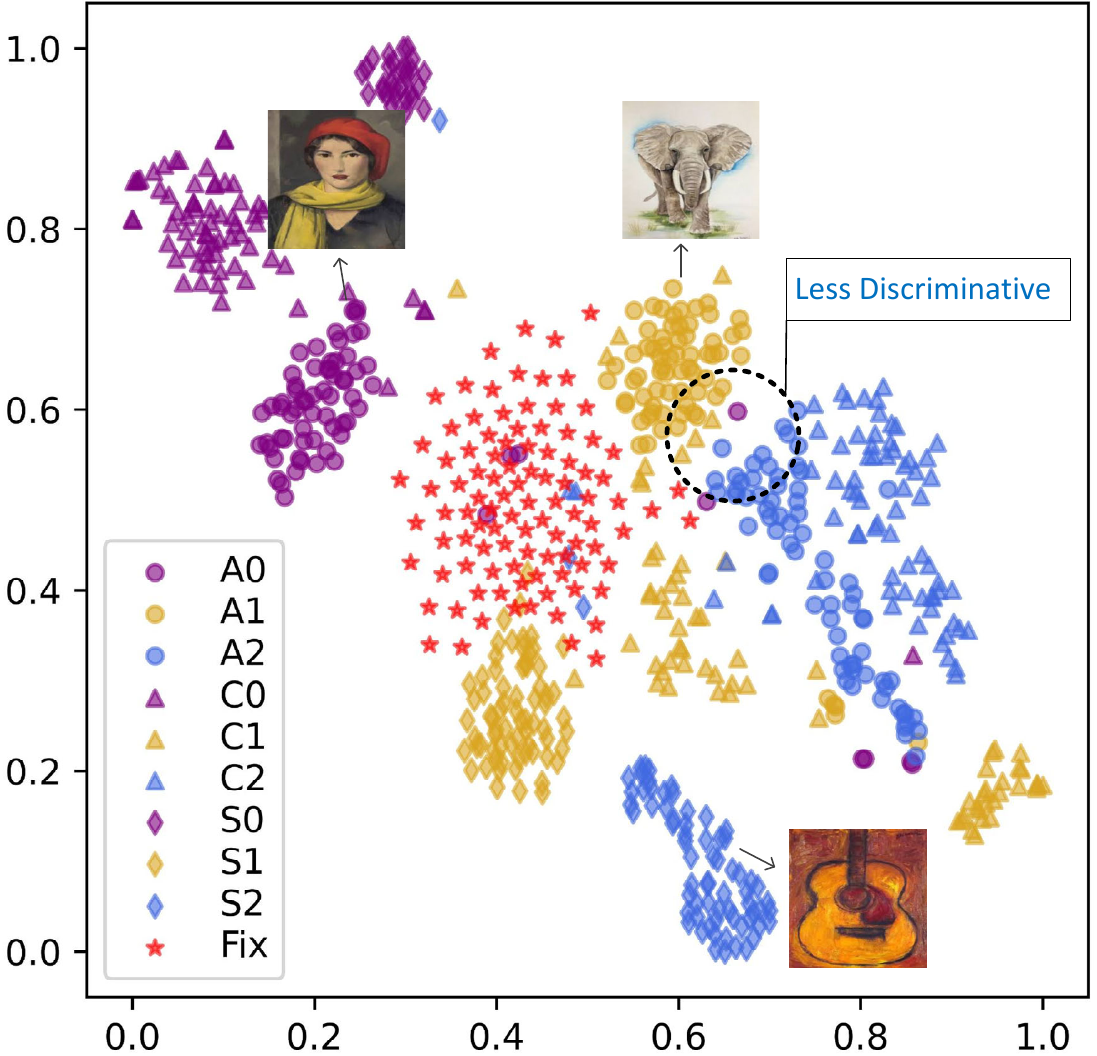}%
\label{fig:VMa}}
\hfil
\subfloat[]{\includegraphics[width=2in]{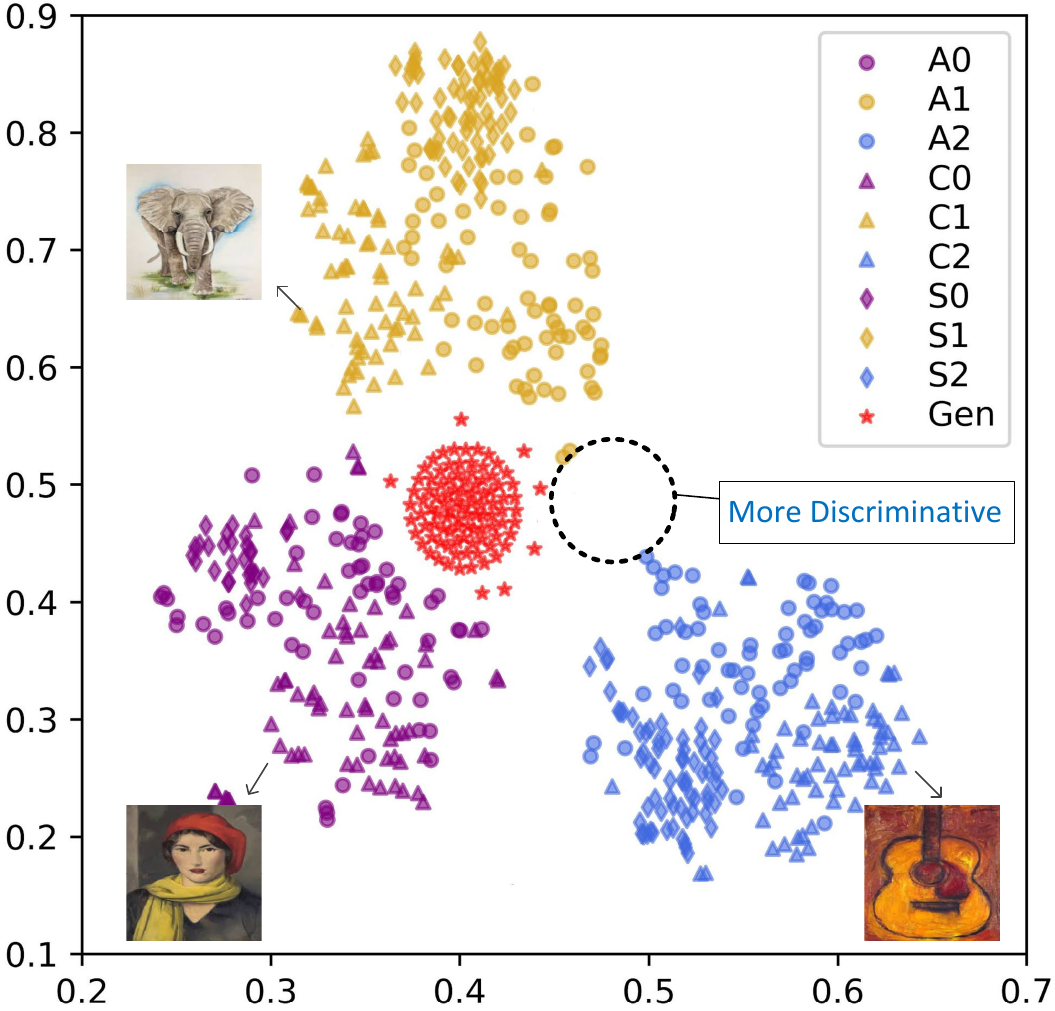}%
\label{fig:VMb}}
\caption{
T-SNE visualization of \mtname features after training (fixed distribution v.s. adaptively generated distribution).
There are three source domains (\textit{A, C, S}) data of the PACS dataset.
In the legend, \textit{A$i$} represents the class label $i$ of source domain \textit{A}.
The meanings of different colors and markers are the same as Fig. \ref{Fig::VisualStart}.
(a) Source domain features and fixed distribution features.
(b) Source domain features and generated distribution features.
}
\label{Fig::VisualMid}
\end{figure*}

\noindent
\textbf{Class-Wise Alignment.} \mtname uses the label information (encoded in a one-hot vector) in the adversarial training.
Thus, the \gename generates features for each class in training.
It means that the distributions of source domains data are aligned in a class-wise manner.
This fine-grained class-wise alignment approach can further improve the performance of \mtname.

\noindent
\textbf{More Discriminative Features.}
Fig. \ref{Fig::VisualMid} shows the source domain features and the reference distribution features after training the model.
The distances between different class clusters in Fig. \ref{fig:VMb} are more evident than that in Fig. \ref{fig:VMa}.
It indicates that \mtname is capable of learning more discriminative features among different classes for different source domains.
Therefore, \mtname has good domain generalization performance.

\section{Experiments}\label{Sec.PE}
In this section, we conduct experiments to evaluate the performance of \mtname.
We first compare \mtname with some recent centralized domain generalization solutions on three different datasets.
Then, we have an ablation study of the \mtname scheme.
Afterward, we investigate the in-domain performance of \mtname.
Last, we study the impact of the different reference distributions.

\subsection{Experimental Settings}
\noindent\textbf{Implementation.} We conduct our experiments using Pytorch 1.7.1 deep learning framework and Python 3.6.5 on Ubuntu 16.04.
We use four Linux terminals to simulate the deployment of \mtname.
Our server uses Geforce RTX 2080ti GPU with 24G RAM for computing.
Following most of the previous studies on FL \cite{Popoola2022FederatedDL,DBLP:journals/iotj/ZhanLQZG20}, we simulate the computation of clients on the Linux server and then measure \mtname performance.
Since the learning process is exactly the same, the performance metrics measured are accurate in our experiments.

\begin{table}
\centering
\renewcommand\arraystretch{\tabarr}
\resizebox{0.38\textwidth}{!}{
  \begin{tabular}{||c|c|c|c||}
  \hline
  datasets ($\rightarrow$) & VLCS & PACS & Office-Home \\
  \hline
  \hline
  $\lambda_0$ & \multicolumn{3}{c||}{0.85} \\
  \hline
  $\lambda_1$ & \multicolumn{3}{c||}{0.15} \\
  \hline
  $lr_f$ & 0.01 & 0.001 & 0.05 \\
  \hline
  $lr_g$ & 0.007 & 0.0007 & 0.007 \\
  \hline
  $lr_d$ & 0.007 & 0.0007 & 0.001 \\
  \hline
 \end{tabular}
}
\caption{Parameters setting in experiments on different datasets.}\label{tb::para-set}
\end{table}

\noindent\textbf{Datasets.}
All experiments are based on three widely used datasets in DG, i.e., VLCS (Pascal \cite{Everingham2009ThePV}, LabelMe \cite{Russell2007LabelMeAD}, Caltech-101 \cite{FeiFei2004LearningGV}, and SUN \cite{Choi2010ExploitingHC}), PACS \cite{8237853} (Photo, Art painting, Cartoon, and Sketch), and Office-Home \cite{DBLP:conf/cvpr/VenkateswaraECP17} (Real-World, Clipart, Product, and Art).
All of them have four sub-datasets, which form distinct domains.
For each dataset, we utilize the leave-one-domain-out validation strategy.
That is, we let one dataset serve as the target domain and use the rest three datasets as source domains.
Like \cite{Ghifary2015DomainGF}, each domain is divided into a training set (70\%) and a validation set (30\%) randomly.
The well-trained model is tested on the target domain data.
Besides, we follow the protocol of \cite{8237853} to perform experiments on PACS.
For Office-Home, we use the same protocol as \cite{DBLP:conf/dagm/DInnocenteC18}.
Since Office-Home and PACS have related domain types, conducting experiments on these two datasets can check the scalability of \mtname when the number of categories varies from 7 to 65.
The three used datasets are standard ones used for studying domain generation.
Thus, we can compare our experimental results with prior solutions.

\noindent\textbf{Network Architecture.}
We test the performance of \mtname by using three pre-trained network architectures as the feature extractor.
The three network architectures are the main structure of AlexNet \cite{Krizhevsky2012ImageNetCW}, ResNet18 \cite{DBLP:conf/cvpr/HeZRS16}, or ResNet50 \cite{DBLP:conf/cvpr/HeZRS16} without including their last layers.
Besides, the classifier consists of the last layer of these pre-trained network structures and an additional output layer.
For \gename and discriminator, both of them have two fully connected layers.
The two layers of \gename and the first layer of discriminator have the same size as the hidden representation.
The size of the second layer in discriminator is set to one.

\noindent\textbf{Model Training.}
When updating the components' parameters in \mtname, each client uses Stochastic Gradient Descent (SGD) to calculate the model gradient.
In order to improve the model expression, the Rectified Linear Unit (ReLU) is used as an activation function to produce non-linear units.
From the data aspect, we use the data augmentation protocol from JiGen \cite{Carlucci2019DomainGB} to improve data quality which can further improve the model expression.
The source domain local epoch $E_0$ (for classification) is 3, $E_1$ (for feature alignment) is 7, and batch size is 16.
The global model with the highest accuracy across all source domains is used to test accuracy on the unseen target domain.

\noindent\textbf{Parameter Settings.}
In all experiments, the parameters of feature extractor are initialized with pre-trained weights using ImageNet \cite{Deng2009ImageNetAL}.
The hyper-parameters of the feature extractor and the initial learning rates of different components on different datasets are detailed in Table \ref{tb::para-set}.
Notice that the hyper-parameters $\lambda_0$ ($0 < \lambda_0 < 1$) and $\lambda_0$ ($0 < \lambda_1 < 1$) train the feature extractor together, and $\lambda_0 + \lambda_1 = 1$.
In particular, $lr_f$ is the learning rate of the feature extractor and classifier, the learning rate of \gename and discriminator are $lr_g$ and $lr_d$, respectively.
In our experiments, unless otherwise stated, the hyper-parameters and learning rates are set as the above default configuration.

\subsection{Performance Evaluation}\label{SEC::EX-Eva}
In this section, we compare \mtname with several recent domain generalization solutions on VLCS, PACS, and Office-Home datasets.
These solutions are briefly introduced as follows.
\begin{enumerate}
  \item
  DANN \cite{ganin2016domain}, a neural network that can both accurately classify source data and have features that are invariant across multiple source domains.
  {DANN is the abbreviation of \textbf{D}omain-\textbf{A}dversarial \textbf{N}eural \textbf{N}etwork.}

  \item
  JiGen \cite{Carlucci2019DomainGB}, a supervised framework for learning to generalize across visual domains by solving jigsaw puzzles.

  \item
 Epi-FCR \cite{li2019episodic}, a scheme to learn domain shift using episodic training.
  {Epi-FCR is the abbreviation of \textbf{Epi}sodic-\textbf{F}eature and \textbf{C}lassifier \textbf{R}egularisation.}

  \item
  MTSSL \cite{albuquerque2020improving}, a method for enabling models to learn transferable features through a self-supervised task of Gabor filter bank response prediction.
  {MTSSL is the abbreviation of \textbf{M}ulti-\textbf{T}ask \textbf{S}elf-\textbf{S}upervised \textbf{L}earning.}

  \item
  EISNet \cite{DBLP:conf/eccv/WangYLFH20}, a network that uses self-supervised learning and metric learning to improve classifier performance on target domains.
  {EISNet is the abbreviation of \textbf{E}xtrinsic and \textbf{I}ntrinsic \textbf{S}upervision \textbf{Net}work.}

  \item
  L2A-OT \cite{DBLP:conf/eccv/ZhouYHX20}, a method to learn domain-invariant features by augmenting the source domain with synthetic data.
  {L2A-OT is the abbreviation of \textbf{L}earning \textbf{to} \textbf{A}ugment by \textbf{O}ptimal \textbf{T}ransport.}
  \item
  DSON \cite{DBLP:conf/eccv/SeoSKKHH20}, a scheme that combines batch normalization and instance normalization to enhance generalization performance on target domains.
  {DSON is the abbreviation of \textbf{D}omain \textbf{S}pecific \textbf{O}ptimized \textbf{N}ormalization.}

  \item
  Mixstyle \cite{DBLP:conf/iclr/ZhouY0X21}, a method for mixing features across source domains to synthesize new source domains to optimize model generalization.

  \item
  RSC \cite{DBLP:conf/eccv/HuangWXH20}, a method to discard dominant features of training data to optimize the generalization ability of a model.
  {RSC is the abbreviation of \textbf{R}epresentation \textbf{S}elf-\textbf{C}hallenging.
  }

\end{enumerate}

All prior solutions require centralized data access, whereas \mtname is used for domain generalization in a distributed way.
We also compare it with a recent state-of-the-art DG method: \FedDG \cite{liu2021feddg}, which does not centralize the dataset and is also trained in the FL setting.
Besides, FedAvg \cite{DBLP:conf/aistats/McMahanMRHA17} is used as a baseline.
We do not compare COPA \cite{WuCollaborativeOA} because of the following reasons.
First, it sacrifices security (refer to Section~\ref{Sec.INTRO} for more details).
Second, the project codes are not publicly available.
We also do not compare FL optimization methods (e.g., FedProx, FedNova, and MOON) since these papers do not focus on the domain generation problem.
These optimization methods differ from \mtname in the following aspects.
First, the source domain data in \mtname are from different clients with domain discrepancy, instead of different subsets from the same dataset \cite{WuCollaborativeOA}.
Second, the discrepancy between the test datasets and training datasets in FedADG also makes it more complex than those in federated optimization methods.
Moreover, \mtname requires building a model that has high performance when testing over the related but unseen target dataset rather than seen dataset.
Note that all solutions used in the comparison are constructed using the same pre-trained network as \mtname.

For each test, we run 5 trails and report the average results which are shown in Table \ref{tb::all-vlcs}, Table \ref{tb::all-pacs}, and Table \ref{tb::all-offho}.
In the three tables, each column containing experimental results (except Avg. column) shows the results when one domain is chosen as the target domain.
We highlight the best results in bold font.

\begin{table*}[ht!]
\footnotesize
  \centering
  \renewcommand\arraystretch{\tabarr}
   \resizebox{\tabWid\textwidth}{!}{
  \begin{tabular}{||c|c|c|cccc|c||}
  \hline
  Paradigm & Backbone & Method & Sun & Pascal & Labelme & Caltech & Avg. \\
  \hline
  \hline
  \multirow{5}*{\shortstack{Centralized w/o \\ privacy concern}} & AlexNet & MTSSL \cite{albuquerque2020improving} & 58.88&62.59&64.99&89.15&67.67\\
  &AlexNet&DANN \cite{ganin2016domain} & 63.60&66.40&64.00&92.60&72.40\\
  &AlexNet&Epi-FCR \cite{li2019episodic} &65.90&67.10&64.30&94.10&72.85\\
  &AlexNet&JiGen \cite{Carlucci2019DomainGB}&64.30&70.62&60.90&96.93&73.19\\
  &AlexNet&RSC \cite{DBLP:conf/eccv/HuangWXH20}&\textbf{68.32}&\textbf{73.93}&\textbf{61.86}&\textbf{97.61}&\textbf{75.43}\\
  \hline
  \hline
  \multirow{2}*{Distributed}&AlexNet&FedAvg \cite{DBLP:conf/aistats/McMahanMRHA17}& 46.65 & 48.77 & 52.32 & 71.43 & 54.79 \\
  &AlexNet&FedADG (ours)&\textbf{71.81}&\textbf{73.40}&\textbf{61.07}&\textbf{93.44}&\textbf{75.09}\\
  \hline
  \hline
  \multirow{2}*{\shortstack{Centralized w/o \\ privacy concern}} & ResNet18 & JiGen \cite{Carlucci2019DomainGB} &71.40&70.93&62.06&96.17&75.14\\[0.8ex]
  &ResNet18&RSC \cite{DBLP:conf/eccv/HuangWXH20}&\textbf{72.10}&\textbf{73.81}&\textbf{62.51}&\textbf{96.21}&\textbf{76.16}\\[0.8ex]
  \hline
  \hline
  \multirow{2}*{Distributed}&ResNet18&FedAvg \cite{DBLP:conf/aistats/McMahanMRHA17}&62.78 & 65.12 & 57.48 & 90.63 & 69.00 \\
  &ResNet18&FedADG (ours)&\textbf{74.95}&\textbf{73.20}&\textbf{61.20}&\textbf{95.78}&\textbf{76.28}\\
  \hline
  \end{tabular}}
\caption{The average classification accuracy using leave-one-domain-out validation on VLCS dataset.}\label{tb::all-vlcs}
\end{table*}

\noindent
\textbf{VLCS.}
Table \ref{tb::all-vlcs} shows the domain generalization accuracy on VLCS.
We use two pre-trained networks, AlexNet and ResNet18, as the backbone to compare \mtname with some recent domain generalization solutions.
Table~\ref{tb::all-vlcs} shows that \mtname outperforms most of the compared centralized solutions.
The performance is comparable to the recent RSC solution.
Besides, \mtname has good performance in both small and large backbone networks.

\noindent
\textbf{PACS.}
Table \ref{tb::all-pacs} shows the domain generalization accuracy on PACS.
We use the same backbone network as in VLCS.
In Table~\ref{tb::all-pacs}, we find that the performance of \mtname is better than most of the compared centralized solutions.
Furthermore, the performance of \mtname is obviously improved compared to \FedDG, which is also a distributed DG method.
Besides, we observe that \FedDG does not improve performance like \mtname (as the backbone size increases from AlexNet to ResNet18).

The accuracy of \mtname is slightly worse than the recent solution L2A-OT.
Specifically, \mtname significantly improves the performance in the Sketch domain.

\begin{table*}[ht!]
 \footnotesize
  \centering
  \renewcommand\arraystretch{\tabarr}
 \resizebox{\tabWid\textwidth}{!}{
  \begin{tabular}{||c|c|c|cccc|c||}
  \hline
  Paradigm & Backbone & Method & Sketch & Artpaint & Cartoon & Photo & Avg. \\
  \hline
  \hline
  \multirow{5}*{\shortstack{Centralized w/o \\ privacy concern}} & AlexNet & DANN \cite{ganin2016domain} & 57.00 & 63.20 & 67.50 & 88.10 & 68.95 \\
  &AlexNet&MTSSL \cite{albuquerque2020improving} & 63.91 & 61.67 & 67.41 & 84.31 & 69.32 \\
  &AlexNet&Epi-FCR \cite{li2019episodic} & 65.00&64.70&72.30&86.10&72.03\\
  &AlexNet&JiGen \cite{Carlucci2019DomainGB}&\textbf{65.18}&\textbf{67.63}&\textbf{71.71}&\textbf{89.00}&\textbf{73.38}\\
  \hline
  \hline
  \multirow{3}*{Distributed}
  &AlexNet&FedAvg \cite{DBLP:conf/aistats/McMahanMRHA17}& 60.52 & 65.97 & 62.93 & 86.95 & 69.09\\
  &AlexNet&\FedDG \cite{liu2021feddg}& 67.63 & 66.50 & 63.51 & \textbf{89.26} & 71.73\\
  &AlexNet&FedADG (ours)&\textbf{69.15}&\textbf{71.68}&\textbf{70.14}&87.01&\textbf{74.50}\\
  \hline
  \hline
  \multirow{5}*{\shortstack{Centralized w/o \\ privacy concern}}
  &ResNet18&Epi-FCR \cite{li2019episodic} & 73.00&82.10&77.00&93.90&81.50\\
  &ResNet18&JiGen \cite{Carlucci2019DomainGB} &71.35&79.42&75.25&96.03&80.51\\
  &ResNet18&EISNet \cite{DBLP:conf/eccv/WangYLFH20} &\textbf{74.33}&81.89&76.44&95.93&82.15\\
  &ResNet18&L2A-OT \cite{DBLP:conf/eccv/ZhouYHX20}&73.60&\textbf{83.30}&\textbf{78.20}&\textbf{96.20}&\textbf{82.81}\\
  \hline
  \hline
  \multirow{3}*{Distributed}
  &ResNet18&FedAvg \cite{DBLP:conf/aistats/McMahanMRHA17}&70.51 & 77.18 & 73.97 & 89.86 & 77.88\\
  &ResNet18&\FedDG \cite{liu2021feddg}& 61.53 & 64.08 & 72.70 & 89.26 & 71.89\\
  &ResNet18&FedADG (ours) &\textbf{78.56}&\textbf{81.39}&\textbf{75.39}&\textbf{93.64}&\textbf{82.25}\\
  \hline
\end{tabular}}
\caption{The average classification accuracy using leave-one-domain-out validation on PACS dataset.}\label{tb::all-pacs}
\end{table*}

\noindent
\textbf{Office-Home.}
We also evaluate \mtname on the Office-Home dataset and the results are shown in Table \ref{tb::all-offho}.
ResNet18 and ResNet50 are applied as the backbone.
In Table~\ref{tb::all-offho}, we observe that \mtname is better than other solutions.

\noindent
\textbf{Compared with Traditional Centralized Method.}
Table \ref{tb::all-vlcs}, Table \ref{tb::all-pacs}, and Table \ref{tb::all-offho} present the domain generalization accuracy of \mtname and prior traditional centralized machine learning solutions without FL.
These traditional ML approaches are described in Section VI-B.
In these tables, the paradigm of these traditional ML solutions is represented as ``centralized w/o privacy concerns''.
That is, these solutions require the centralized server to access source domain data and expose sensitive local information.
In Table \ref{tb::all-vlcs} and Table \ref{tb::all-pacs}, we find that the performance of \mtname is comparable to the traditional ML methods.
In particular, the generalization accuracy of the AlexNet-based \mtname on PACS is over 1\% higher than the centralized approaches (e.g., MTSSL, Epi-FCR, JiGen).
Besides, Table \ref{tb::all-offho} shows that \mtname performs significantly better than other traditional ML methods on Office-Home dataset.
In summary, our proposed \mtname can achieve good domain generalization capability while still protecting data privacy.
\begin{table*}[ht!]
\footnotesize
  \centering
\renewcommand\arraystretch{\tabarr}
 \resizebox{\tabWid\textwidth}{!}{
  \begin{tabular}{||c|c|c|cccc|c||}
  \hline
  Paradigm & Backbone & Method & Real & Clipart & Product & Art & Avg. \\
  \hline
  \hline
  \multirow{3}*{\shortstack{Centralized w/o \\ privacy concern}} & ResNet18 & JiGen \cite{Carlucci2019DomainGB} &72.79&47.51&71.47&53.04&61.20\\
  &ResNet18&DSON \cite{DBLP:conf/eccv/SeoSKKHH20} & \textbf{74.68}&45.70&\textbf{71.84}&\textbf{59.37}&62.90\\
  &ResNet18&RSC \cite{DBLP:conf/eccv/HuangWXH20}&74.54&\textbf{47.90}&71.63&58.42&\textbf{63.12}\\
  \hline
  \hline
  \multirow{2}*{Distributed}&ResNet18&FedAvg \cite{DBLP:conf/aistats/McMahanMRHA17}& 71.31 & 52.11 & 67.60 & 48.00 & 59.76\\
  &ResNet18&FedADG (ours)&\textbf{74.98}&\textbf{53.98}&\textbf{70.83}&\textbf{58.13}&\textbf{64.48}\\
  \hline
  \hline
  \multirow{2}*{\shortstack{Centralized w/o \\ privacy concern}} & ResNet50 & Mixstyle \cite{DBLP:conf/iclr/ZhouY0X21} & 69.20&\textbf{53.20}&68.20&51.10&60.43\\[0.8ex]
  &ResNet50&RSC \cite{DBLP:conf/eccv/HuangWXH20}&\textbf{75.10}&51.40&\textbf{74.80}&\textbf{60.70}&\textbf{65.50}\\[0.8ex]
  \hline
  \hline
  \multirow{2}*{Distributed}&ResNet50&FedAvg \cite{DBLP:conf/aistats/McMahanMRHA17}& 71.83 & 54.06 & 69.14 & 53.07 & 62.03 \\
  &ResNet50&FedADG (ours)&\textbf{76.48}&\textbf{56.09}&\textbf{74.87}&\textbf{60.27}&\textbf{66.93}\\
  \hline
  \end{tabular}}
\caption{The average classification accuracy using leave-one-domain-out validation on Office-Home dataset.}\label{tb::all-offho}
\end{table*}

\begin{table*}[ht!]
  \footnotesize
  \centering
  \renewcommand\arraystretch{\tabarr}
  \resizebox{0.75\textwidth}{!}{
  \begin{tabular}{||c|c|c|cccc|c||}
  \hline
  Paradigm&PACS &Backbone& Sketch & Artpaint & Cartoon & Photo & Avg. \\
  \hline
  \hline
  \multirow{4}*{Distributed}&FedAvg (in)&AlexNet&\textbf{99.98}&\textbf{99.98}&\textbf{99.97}&\textbf{99.98}&\textbf{99.98}\\
  &FedADG (in)&AlexNet&98.13&98.06&97.96&98.03&98.05\\
  \cline{2-8}

  &FedAvg (in)&ResNet18&99.12&98.08&98.38&95.83&97.85\\
  &FedADG (in)&ResNet18&\textbf{99.83}&\textbf{98.47}&\textbf{94.08}&\textbf{99.73}&\textbf{98.03}\\
  \hline
  \end{tabular}
  }
\caption{The average in-domain classification accuracy using leave-one-domain-out validation on PACS dataset.}\label{tb::all-fedIn}
\end{table*}

\subsection{Ablation Study}
\begin{figure}[ht]
\centering
\includegraphics[width=3in]{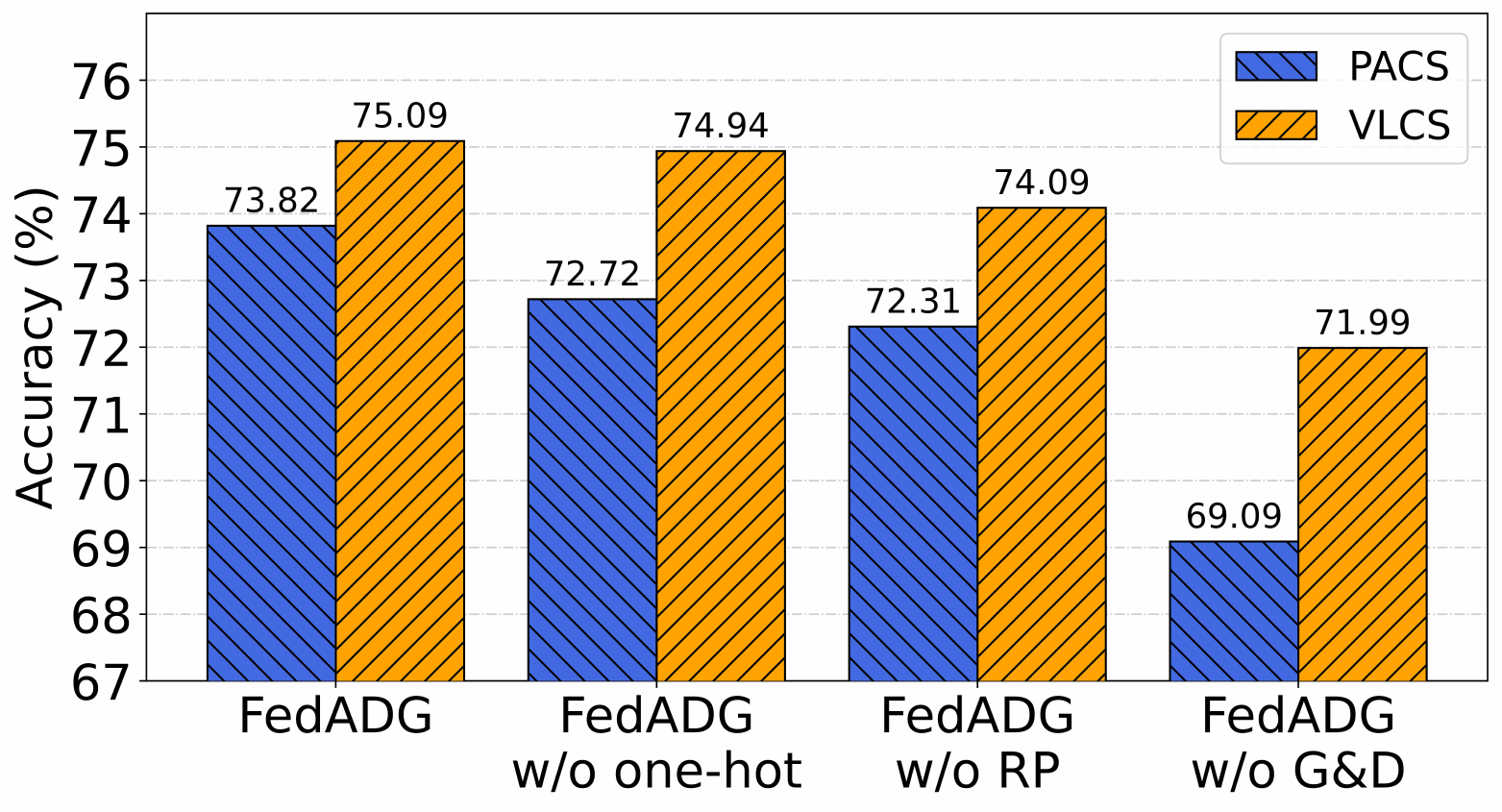}
\caption{
Ablation study on VLCS and PACS datasets.
``\textit{\mtname w/o one-hot}'' means to remove the one-hot vector from \mtname.
``\textit{\mtname w/o RP}" represents the model by removing the random projection layer from \mtname.
Besides, ``\textit{FedADG w/o G\&D}" represents the model by removing \gename and discriminator from \mtname.
  }
\label{Fig::abl}
\end{figure}

An ablation study investigates the performance of \mtname by removing a certain component to
understand the contribution of the component to the overall \mtname scheme.
We perform ablation experiments on VLCS and PACS datasets using AlexNet.
Specifically, we focus on the distribution generator and discriminator, along with the data label (encoding as a one-hot vector) in these two components.
When we remove the one-hot vectors from both distribution generator and discriminator, the remained \mtname is denoted as ``\mtname w/o one-hot''.
We use ``\mtname w/o RP'' to denote \mtname without the Random Projection (RP) layer.
``\mtname w/o G\&D'' represents \mtname without \gename and discriminator.
Fig. \ref{Fig::abl} shows the ablation study results.
The results analysis is performed as follows.

\noindent\textbf{\mtname w/o one-hot.}
As shown in Fig. \ref{Fig::abl},
\mtname has higher accuracy in each target domain than \mtname w/o one-hot.
These results demonstrate that the class-wise alignment can increase the generalization performance of \mtname.

\noindent\textbf{\mtname w/o RP.}
Fig. \ref{Fig::abl} shows that the accuracy of \mtname w/o RP is less than \mtname.
The function of random projection is to decrease the dimension of features.
The low-dimension features stabilize the training of ALN and help to do feature alignment.
Thus, the random projection layer can improve the performance of \mtname on target domains.

\noindent\textbf{FedADG w/o G\&D.}
Fig. \ref{Fig::abl} shows that the accuracy of FedADG w/o G\&D is less than \mtname.
In terms of average accuracy, \mtname
is over 3\% higher than FedADG w/o G\&D because the latter lacks of domain generalization design.

\subsection{In-domain Performance Evaluation}
The previous experimental results are tested on out-of-domain data.
We also measure the performance of \mtname on in-domain data.
We consider the commonly used experimental setting: both the training and testing data come from the same domain.
AlexNet and ResNet18 are applied as the backbone and PACS dataset is used in experiments.
Table \ref{tb::all-fedIn} compares the in-domain performance of \mtname and FedAvg.
In Table \ref{tb::all-fedIn}, ``FedAvg (in)'' and ``\mtname (in)'' represent the in-domain performances of FedAvg and \mtname, respectively.

The results show that the in-domain performance of \mtname is comparable to that of FedAvg.
It indicates that \mtname can be used on both source domains and the unseen target domain.
In practice, the clients can train both FedAvg (used for in-domain data) and \mtname (used for out-domain data).

\subsection{Impact of Different Reference Distributions}
As we have discussed in Section~\ref{Sec.Components}, the reference distribution helps to align the feature distributions of all source domain data, which in turn improves the accuracy of classification on target domains.
In general, most existing works adopt fixed reference distribution without considering the distortion it may cause to the source domain distribution.
In this part, we investigate how the adaptively generated distribution can outperform the fixed settings such as Gaussian distribution ($\mathcal{N}$), Uniform distribution ($\mathcal{U}$), and Laplace distribution \cite{li2018domain}.
The experimental results of different reference distributions using AlexNet on PACS are shown in Table \ref{tb::refe-pacs}.

\begin{table}
\centering
\renewcommand\arraystretch{\tabarr}
\resizebox{0.48\textwidth}{!}{
  \begin{tabular}{||c|cccc|c||}
  \hline
  Unseen domain ($\rightarrow$) & Sketch & Artpaint & Cartoon & Photo & Avg.\\
  \hline
  \hline
  \multicolumn{6}{|c|}{fixed reference distribution} \\
  \hline
  $\mathcal{N}\sim(0,\mathbf{I})$ & 49.45 & 53.32 & 53.54 &75.69 & 58.00 \\
  $\mathcal{U}\sim[-1,1]$ & 38.23 & 57.71 & 55.33 & 81.26 & 58.13 \\
  Laplace$(1/\sqrt{2})$ & 44.29 & 54.69 & 55.84 & 84.31 & 59.78 \\
  \hline
  \hline
  \multicolumn{6}{|c|}{adaptively generated distribution} \\
  \hline
  \mtname (ours) & \textbf{69.15} & \textbf{68.99} & \textbf{70.14} & \textbf{87.01} & \textbf{73.82} \\
  \hline
 \end{tabular}}
\caption{Experimental results when using different reference distributions in \mtname on PACS dataset.}\label{tb::refe-pacs}
\end{table}

In Table~\ref{tb::refe-pacs}, the parameter of the Laplace distribution we compared in the experiment is $1/\sqrt{2}$, which is proved by  Li \emph{et al.} \cite{li2018domain} to have the best effect on target domains.
Moreover, we find that the accuracy of the Laplace distribution with the parameter of $1/\sqrt{2}$ is higher than the other two fixed distributions in the table.
By observing all experimental results in Table~\ref{tb::refe-pacs}, we notice that the average accuracy of the adaptively generated reference distribution can be 10\% higher than the accuracy of the fixed reference distributions.
Especially in the target domains of Cartoon and Sketch, the accuracy of using the adaptively generated distribution in \mtname is about 20\% higher than the accuracy of the fixed reference distribution.
The remarkable result of \mtname supports the effectiveness of using adaptively generated distribution.
It proves that the generated adaptive reference distribution can promote the performance of the model for target domains.

\section{Limitation}\label{Sec.Limit}
The major limitation of \mtname is that it requires extra resources consumption in training compared with the traditional FL schemes (e.g., FedAvg).
On the one hand, clients need more computing resources to perform local adversarial training of ALN.
On the other hand, clients have a larger communication overhead since they are required to submit the updates of ALN to the server as well as receive a new global ALN model in federated training.
In a nutshell, \mtname achieves DG at the cost of extra resources consumption.
Note that the extra resources consumption is not very heavy and can be easily handled by the majority of smart devices on the market.

\section{Conclusion}\label{Sec.Conc}
In this paper, we propose \mtname scheme under the federated learning setting for domain generalization.
The main idea of \mtname is to learn the domain-invariant feature representation in FL while eliminating the requirement for a centralized server to access clients’ local data.
First, we propose the federated adversarial learning approach to measure and align the distributions among different source domains via matching each distribution to the reference distribution.
Specifically, we use the federated adversarial learning technique to adaptively learn a dynamic distribution (by accommodating all source domains) as the reference distribution.
Therefore, the learned feature representation tends to be universal.
Then, our proposed \mtname uses the adaptively generated reference distributions and class-wise alignment technique.
It ensures that \mtname has good generalization performance over the unseen target domains while protecting local data privacy.
Furthermore, we analyze the explainability of \mtname, which helps researchers to optimize the model and make the model more trustworthy.
Finally, the effectiveness of \mtname has been demonstrated by intensive simulations.
Thus, \mtname significantly boosts FL performance.
There are two directions to launch further research.
First, we aim to handle the scenario in which the unseen target domain contains more classes than the seen source domain.
Second, we plan to find an optimization method to automatically balance the classification training epoch and the alignment training epoch to obtain a better federated generalization performance.


{\small
\bibliographystyle{ieee_fullname}
\bibliography{bibfile}
}

\end{document}